\pdfoutput=1

\documentclass[11pt]{article}

\usepackage{acl}

\usepackage{times}
\usepackage{latexsym}
\usepackage{todonotes}
\usepackage[T1]{fontenc}
\usepackage{amsfonts}
\usepackage{tipa}
\usepackage{microtype}

\usepackage{pifont}
\usepackage{algorithm}
\usepackage{enumitem}
\usepackage{algorithmic}
\usepackage{graphicx}
\usepackage{subfigure}
\usepackage{multirow}
\usepackage{makecell}
\usepackage{balance}
\usepackage{booktabs}
\usepackage{soul}
\title{OpenSR: Open-Modality Speech Recognition \\
via Maintaining Multi-Modality Alignment}

\author{
Xize Cheng\textsuperscript{\rm 1}\Thanks{ Equal contribution.} ,
Tao Jin\textsuperscript{\rm 1*} ,
Linjun Li\textsuperscript{\rm 1*} ,
Wang Lin\textsuperscript{\rm 1} ,
Xinyu Duan\textsuperscript{\rm 2} ,
Zhou Zhao\textsuperscript{\rm 1} \Thanks{ Corresponding author.}\\
\small\textsuperscript{\rm 1}Zhejiang University ,\small\textsuperscript{\rm 2}Huawei Cloud\\
\tt\small \{chengxize,jint\_zju,zhaozhou\}@zju.edu.cn 
\tt\small duanxinyu@huawei.com
}

\begin{document}
\maketitle

\begin{abstract}

Speech Recognition builds a bridge between the multimedia streaming (audio-only, visual-only or audio-visual) and the corresponding text transcription. However, when training the specific model of new domain, it often gets stuck in the lack of new-domain utterances, especially the labeled visual utterances. 
To break through this restriction, we attempt to achieve zero-shot modality transfer by maintaining the multi-modality alignment in phoneme space learned with unlabeled multimedia utterances in the high resource domain during the pre-training \cite{shi2022learning}, and propose a training system Open-modality Speech Recognition (\textbf{OpenSR}) that enables the models trained on a single modality (e.g., audio-only) applicable to more modalities (e.g., visual-only and audio-visual).
Furthermore, we employ a cluster-based prompt tuning strategy to handle the domain shift for the scenarios with only common words in the new domain utterances.
We demonstrate that OpenSR enables modality transfer from one to any in three different settings (zero-, few- and full-shot), and achieves highly competitive zero-shot performance compared to the existing few-shot and full-shot lip-reading methods.
To the best of our knowledge, OpenSR achieves the state-of-the-art performance of word error rate in LRS2 on audio-visual speech recognition and lip-reading with 2.7\% and 25.0\%, respectively. The code and demo are available at \url{https://github.com/Exgc/OpenSR}.


\end{abstract}

\section{Introduction}


Speech Recognition \cite{afouras2018deep,ren2021learning,zhao2020hearing} (e.g., Audio-Visual Speech Recognition known as AVSR) transcribs visual and audio data into text form, building a bridge between multi-media speech~\citep{cheng2023mixspeech,huang2023make,huang2023audiogpt,cui2022varietysound,GASLT} and natural language~\citep{yin2022mlslt,yin2021simulslt,jin2022mc,jin2022prior,jin2021contrastive}. Among them, ASR (Automatic Speech Recognition) and lip-reading (VSR, Visual Speech Recognition) are twin tasks transcribed using only audio and only vision, respectively. 
Audio utterances with clear pronunciation are sufficient for ASR training, most of which can be easily collected from recordings of telephone conversations and audiobooks \cite{korvas_2014}.
While current lip-reading training systems require mostly-frontal and high-resolution videos with a sufficiently high frame rate, such that motions around the lip area are clearly captured \cite{prajwal2022sub}. 
The significant difficulty of collecting labeled visual utterances hinders the training of lip-reading models suitable for new domains or low-resource domains, resulting in the relatively low-speed development of domain-specific lip-reading models, compared with ASR models.

\begin{figure}[tb]
    \centering
    \includegraphics[scale=0.3]{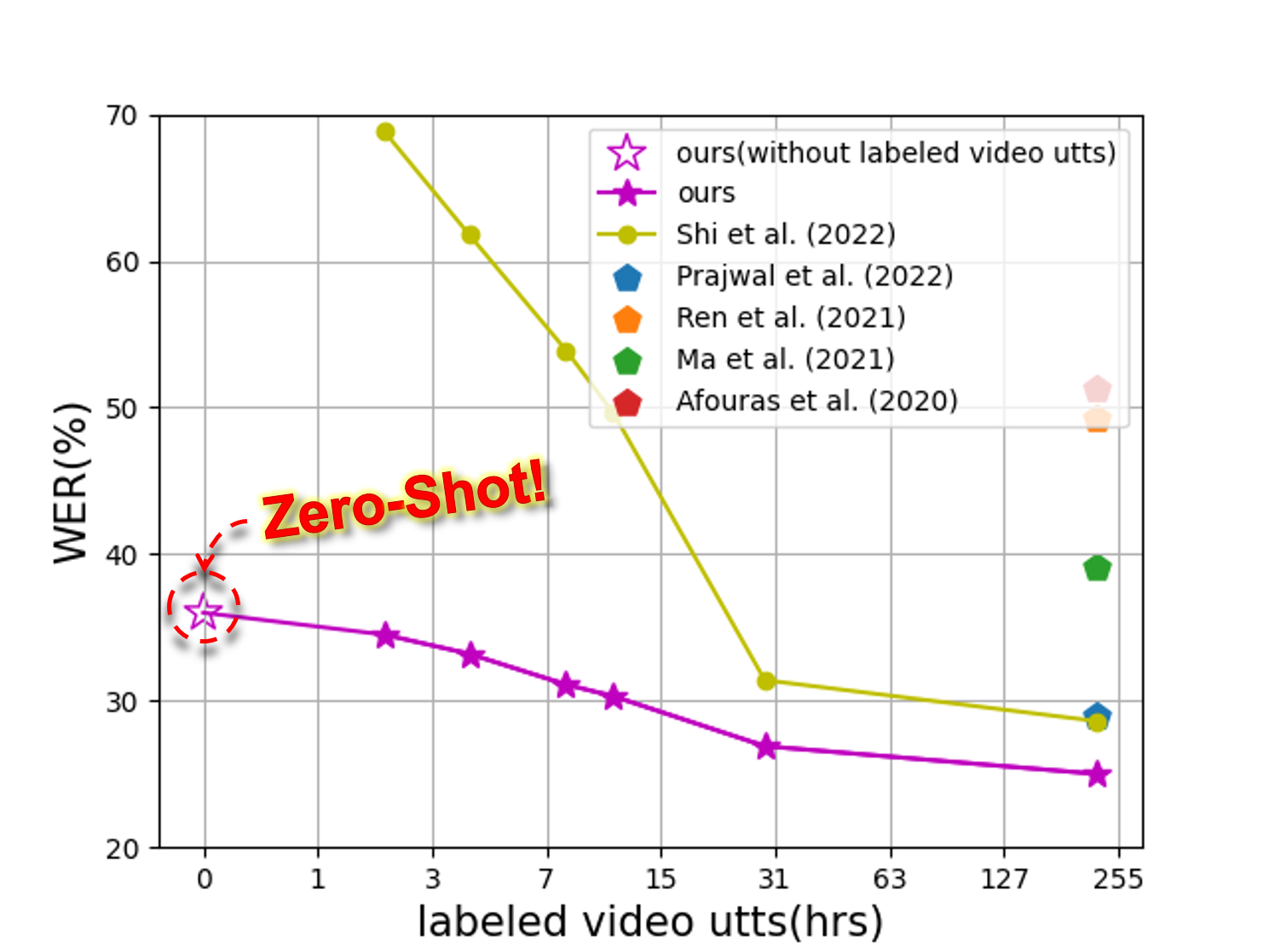}
    \caption{Comparison between OpenSR and previous methods under different sizes of labeled visual utterance. As the results highlighted by the red circle show, OpenSR with zero-shot even outperforms most previous methods with full-shot.}
    \label{fig:zero-shot}
    \vspace*{-0.6cm}
\end{figure}

Since audio speech is easier to obtain, can we use audio utterance alone to train lip-reading models for target domains? Humans, once they have mastered a language, can immediately predict the lip movements of a phoneme \cite{meltzoff1977imitation}. Similarly, if we align the phoneme representations of acoustic and image fragments with each other, we can then apply the term distribution and syntax specification of the target domain learned from the audio utterance to lip-reading. 
Building on this novel idea, we employ the audio-visual aligned encoder, such as AV-Hubert, co-trained on a large number of multi-modality high resource domain utterances, to align the multi-modality utterances in the same phoneme space, and train a domain-specific decoder from phoneme space to text using the labeled audio-only utterances.
Since the audio-visual encoder is trained to embed different modalities of the same phoneme near one another (for instance, the visual phoneme feature of \textrm{/tu\textipa{:}/} is aligned with its audio phoneme feature), we can flexibly transfer the knowledge of the target domain (e.g., the mapping from phoneme \textrm{/tu\textipa{:}/} to homophones \texttt{two} and \texttt{too}, and the syntax used to distinguish them) from a single modality (i.e., audio) to more modalities (i.e., visual-only and audio-visual). As shown in Figure \ref{fig:zero-shot}, the zero-shot lip-reading performance of our method is even better than most full-shot methods.




Furthermore, we propose to enhance the discrimination of visually similar words (e.g., \texttt{pet} and \texttt{bet}, which are audio distinguishable but visually similar \cite{Kim2022DistinguishingHU}) with more readily available utterances containing only common words. 
However, the uneven word distribution causes the sequence modeling to be easily corrupted by domain shift during fine-tuning \cite{huang2022generspeech}. 
We propose a cluster-based prompt tuning strategy, \texttt{Cluster Prompt}, only tuning the limited parameters of the fine-grained prompt-embedding of each visual phoneme cluster, to prevent excessive impact on sequence modeling.



The main contributions are as follows: 
\begin{itemize}[itemsep=-4pt,topsep = 0 pt]
    \item To the best of our knowledge, OpenSR is the first to achieve zero-shot lip-reading, which fully considers the modality transfer.
    \item OpenSR is the first low-resource lip-reading method that attempts to leverage common words, adopting the \texttt{Cluster Prompt} strategy to overcome the ensuing domain shift problem and improve the accuracy of lip-reading by 19.1\% to 36.1\%.
    \item OpenSR achieves modality transfer and the state-of-the-art performance in all three settings (zero-shot, few-shot and full-shot). In particular, it achieves 2.7\% and 25\% word error rate on LRS2 in AVSR and lip-reading.
\end{itemize}

\section{Related Work}
\subsection{Lip Reading}
The lip-reading task has attracted many researchers \cite{cooke2006audio,afouras2018deep}, aiming to recognize spoken sentences according to the given video of lip movements without relying on the audio stream.
With the support of a large amount of visual speech utterances \cite{afouras2018deep,afouras2018lrs3}, \citet{assael2016lipnet} first proposes to use neural networks for lip reading. \cite{ma2021lira} and \cite{shi2022learning} adopt different pre-training strategies, attempting to obtain fine-grained lip representation by pre-training on a large amount of additional unlabeled audio-visual utterances \cite{Chung2018VoxCeleb2DS}. 
Some works \cite{Makino2019RecurrentNN,Serdyuk2021AudioVisualSR} adopt massive labeled visual utterances (even more than 90,000h) for training and promote the generalization of the model. However, none of these methods can train lip-reading models without large amounts of labeled visual utterances, making them unusable in low-resource domains where labeled visual utterances are scarce or unavailable. This is also the original intention of our work, for which we propose a training system that can employ only labeled audio utterances to train the lip-reading model.


\subsection{Transfer Learning from Audio to Video}

Although the two parallel speech modalities, audio and video, remain aligned at the temporal level, ASR is still far more accurate than lip reading, benefiting from its easy access to labeled audio utterance and fine-grained phoneme audio representation. \citet{ma2021lira} hopes to take advantage of this natural temporal alignment and use audio to assist with lip-reading training. \citet{ren2021learning} proposes different distillation strategies (from ASR to lip-reading) which enables lip-reading to learn complementary and discriminant clues from ASR. 
\citet{shi2022learning} has adopted audio assisted pre-training methods, that regard audio as the auxiliary supervision for visual utterances in order to obtain fine-grained phoneme visual representations. 
However, in previous methods~\cite{ren2021learning,huang2021multi}, audio only played the role of auxiliary supervision, and the lip-reading models could not be trained using non-visual utterances (i.e., audio) alone. In our work, we attempt to maintain the alignment of audio and video in phoneme space, and employ only labeled audio utterances instead of visual utterances to train the lip-reading model.


\begin{figure*}[tb]
    \centering
    \includegraphics[scale=0.4]{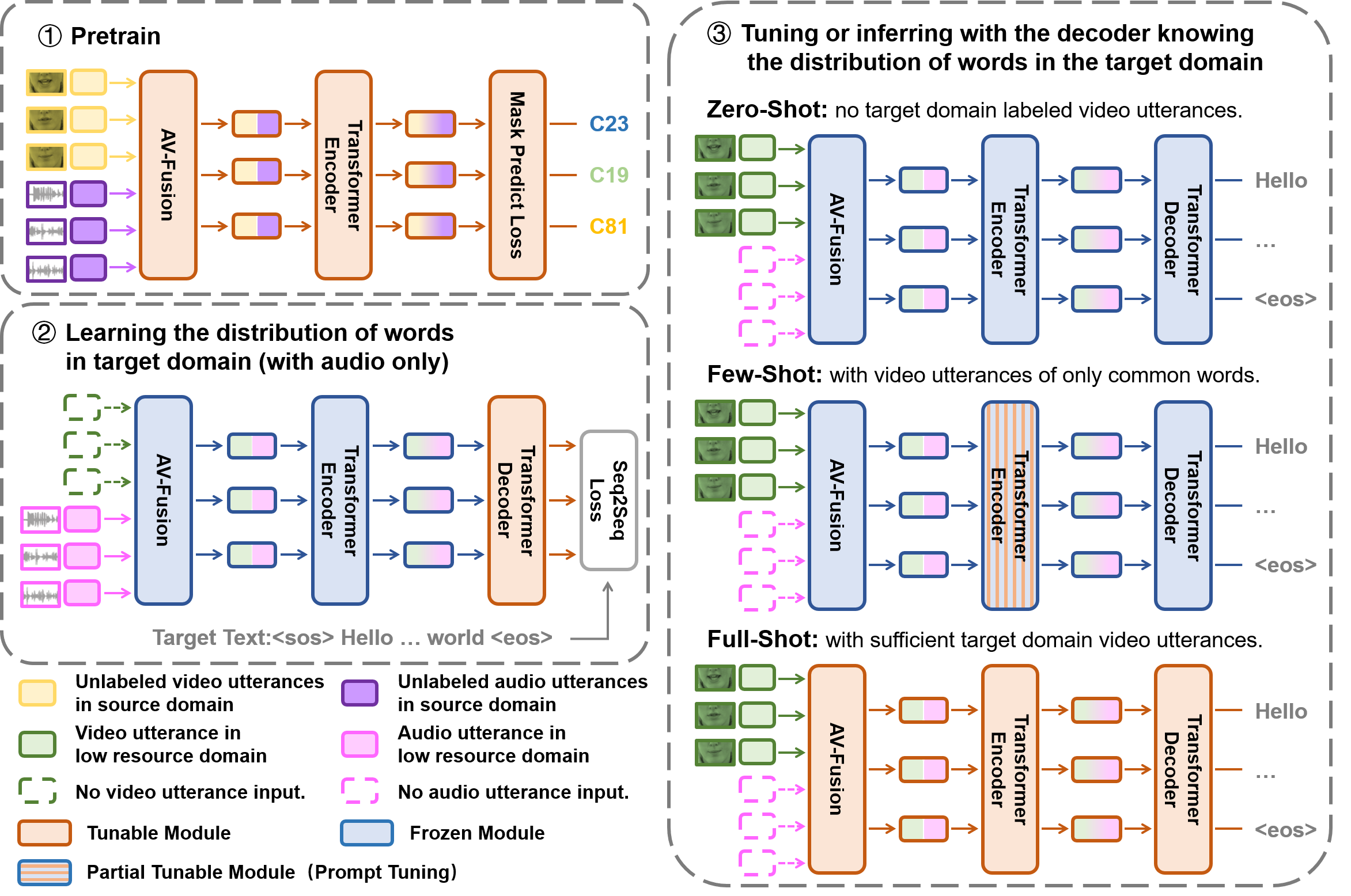}
    \caption{Illustration of OpenSR training system. In the second stage, parameters other than those in the transformer decoder are frozen to maintain the cross modality alignment state achieved in the first stage. We propose three different tuning and inference strategies in the third stage for scenarios with different scales of labeled visual utterances in the target domain.}
    \label{fig:system}
\end{figure*}

\section{Method}

\subsection{Audio-Visual Alignment Learning}
AV-Hubert \cite{shi2022learning} is a self-supervised representation learning method for audio-visual speech, alternating between feature clustering and mask prediction, as shown in the first stage of Figure \ref{fig:system}. 
During the feature clustering, the audio-visual speech is labeled as a sequence of frame-level assignments $z\texttt{=}{\{z^a_t\}}^T_{t\texttt{=}1}$ based on the sequence of image sequences $V\texttt{=}{\{V_t\}}^{T}_{t\texttt{=}1}$ and audio acoustic frames $A\texttt{=}{\{A_t\}}^{T}_{t\texttt{=}1}$ (MFCC or audio-visual features from the previous encoder) with a discrete latent variable model (e.g., k-means). Subsequently, with the paired data ($A, V, z$), the model learns a better audio-visual representation in phoneme space $f^{p}\texttt{=}{\{f^{p}_t\}}^T_{t\texttt{=}1}\in\mathbb{R}^{T \times D}$, where $T$ is the length of the sequence and $D$ is the dimension of the embedding, by reducing the mask prediction loss, just like the mask language modeling in BERT \cite{devlin2018bert}. The above two steps are repeated during training to improve the quality of audio-visual speech clustering and representation.

Furthermore, the random dropout of modalities in the framework maps the speech features of different modalities into the same phoneme space, which not only optimizes the representation quality of the uni-modality, but also achieves cross-modality representation alignment with a large amount of source domain (i.e., high resource domain) unlabeled audio-visual utterances. OpenSR is the first attempt to take full advantage of this cross-modality alignment of speech representations.

\subsection{Decoder Training with Audio Only}
Now that the audio and visual representations have been mapped into the same phoneme space, we can use audio of the target domain as an alternative to video when labeled visual utterance is scarce.

As shown in the second stage in Figure \ref{fig:system}, we adopt the AV-Fusion and the pre-trained Transformer Encoder in the first stage to obtain the features of phoneme space $f^p\texttt{=}{\{f^p_t\}}^T_{t\texttt{=}1}\in\mathbb{R}^{T \times D}$. With only labeled audio utterance of the target domain $f^a\texttt{=}{\{f^a_t\}}^T_{t=1}\in\mathbb{R}^{T \times D}$ as input, the audio-visual feature fed into AV-Fusion can be formally expressed as: $f^{av}_t\texttt{=concat}(f^a_t,\mathbf{0}_{D})\in\mathbb{R}^{T \times 2D}$, similar to the modality dropout mechanism in pre-training stage. 
With the parameters of AV-Fusion and the Transformer Encoder being frozen, we obtain the fusion features $f^m\texttt{=}{\{f^m_t\}}^T_{t\texttt{=}1}\in\mathbb{R}^{T \times D}$:
\begin{equation}
    f^m=\texttt{AV-Fusion}_{(\textcolor{blue}{frozen})}(f^{av})
\end{equation}
and then encode them into a common phoneme space, 
\begin{equation}
    f^p=\texttt{encoder}_{(\textcolor{blue}{frozen})}(f^{m})
\end{equation}
The freezing parameter allows speech recognition training to focus on the decoder (from the phoneme space to the text space), while avoiding the missing visual modality utterances from destroying the cross-modality alignment relationship in the phoneme space. A tunable Transformer Decoder is appended to autoregressively decode the phoneme feature $f^p$ into the target probabilities:
\begin{equation}
    p(w_t|{\{w_i\}}^{t-1}_{i=1}, f^p)\texttt{=decoder}_{(\textcolor{red}{tunable})}(f^{p})
\end{equation}
, where $\{w_i\}_{i\texttt{=}1}^s$ is the ground-truth transcription. In the second and third stages of training, the overall model is trained with cross-entropy loss  $L_{s2s}\texttt{=}-\sum_{t=1}^s{\log{p(w_t|{\{w_i\}}^{t-1}_{i=1}, f^p)}}$.


\subsection{Tuning of Target-domain Decoder in Lip-reading}

The decoder trained with the target domain audio utterances fully learns the word distribution and syntactic characteristics of the target domain. Furthermore, the OpenSR training system can further tune the lip-reading model with the labeled visual utterances. Depending on the amount of visual utterances, it can be divided into three types: zero-shot, few-shot and full-shot.
 
\paragraph{Zero-Shot} In OpenSR, the target domain decoder trained in the ASR can be directly transferred to the target domain lip-reading. Both audio and video are aligned in the same phoneme space, and the co-usable decoder for multi-modality speech recognition only needs to map from phoneme space to text. In inferring on lip-reading, only the visual utterances $f^v\texttt{=}{\{f^v_t\}}^T_{t\texttt{=}1}\in\mathbb{R}^{T \times D}$ inputs into AV-Fusion, and the fusion feature can be formally expressed as: $f^{av}_t\texttt{=concat}(\mathbf{0}_{D},f^v_t)$. 


\paragraph{Full-Shot}
When sufficient labeled visual utterance is available for training, the model parameters can be further fine-tuned. 
With the word distribution and specific syntax of the target domain, the decoder can steadily boost the discrimination and representation of the pre-trained encoder for visual utterance.

\begin{figure}[tb]
    \centering
    \includegraphics[scale=0.5]{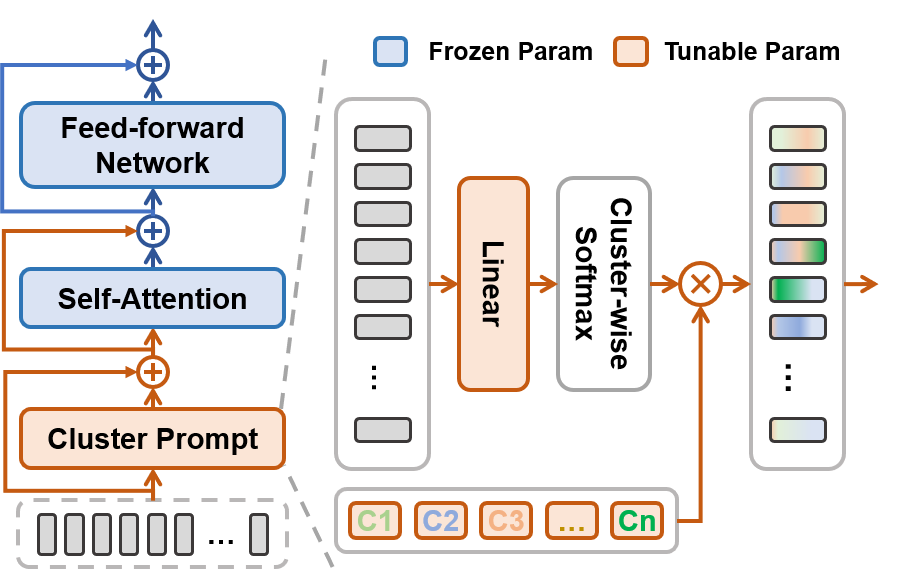}
    \caption{Illustration of Cluster Prompt. The cluster-wise softmax enables the model to assign appropriate prompt embeddings to features based on clustering.}
    \label{fig:cluster}
\end{figure}

\paragraph{Few-Shot}
More commonly, when we can only use visual utterances containing only common words for training, the training of the model is likely to be affected by the data distribution, and the model will be more inclined to recognize common words. We adopt a prompt tuning strategy called \texttt{Cluster Prompt} to make the model pay more attention to local phoneme features, as shown in the Figure \ref{fig:cluster}.

In the first stage, the pre-training process uses $k$-means to put cluster labels on audio-visual features during training. We further explore this cluster-based approach by tuning the learnable clustering embeddings $\mathbf{c}_j\texttt{=}{\{\mathbf{c}_j^i\}}^N_{i\texttt{=}1}\in\mathbb{R}^{N \times D}$ for each cluster in each layer $\texttt{layer}_j$ of the encoder, where $N$ is the number of the clusters. 
Cluster embedding is the cluster-specific fine-grained bias from audio-based to visual-based phoneme features, which is used to further enhance the visual discriminability between visually similar clusters.
The input $\mathbf{x}_j\texttt{=}{\{\mathbf{x}_j^t\}}^T_{t=1}\in\mathbb{R}^{T \times D}$ to each layer is first fed into a cluster network $\texttt{Meta}$, which is consisted of a linear projection layer and a cluster-wise softmax layer, to extract the cluster weights $\mathbf{u}_j\in\mathbb{R}^{T \times N}$ for each phoneme features $\mathbf{x}_j$:
\begin{equation}
    \mathbf{u}_j=\texttt{Meta}_{(\textcolor{red}{tunable})}(\mathbf{x}_j)
\end{equation}
and then combine the cluster embedding vector to update each audio-visual fusion feature:
\begin{equation}
    \mathbf{x'}_j= \mathbf{x}_j+ \mathbf{u}_j\times \mathbf{c}_{j(\textcolor{red}{tunable})}
\end{equation}

At the same time, we freeze the parameters of the encoder and the decoder to maintain the learned syntactic knowledge and reduce the tuning computational resources:
\begin{equation}
    \mathbf{x}_{j+1}=\texttt{layer}_{j(\textcolor{blue}{frozen})}(\mathbf{x'}_j)
\end{equation}


 
\begin{table*}[tb]
\centering
\caption{Comparison of LRS2-COMMON with LRS2: TF stands for word frequency threshold, red numbers indicate the number of utterances containing words that do not appear in the corresponding vocab.}
\begin{tabular}{lrllrrl}
\toprule
\textbf{Split-Name}                                    & \textbf{Train} & \textbf{Test}                       & \textbf{Val}                        & \textbf{Vocab-Size} & \textbf{Hours} & \textbf{TF}      \\ 
\midrule
$\textsc{LRS2-224H}$                                   & 142\,157    & $1\,082$                                & $1\,243$                                &     41\,370                & 224h           & -                \\
$\textsc{LRS2-29H}$                                    & 45\,839          & $1\,082$                                & $1\,243$                                & 17\,660               & 29h            & -                \\
$\textsc{LRS2-Common}_{\scriptsize\texttt{TF\textgreater   10}}$  & 22\,669          & $1\,082_{(514+\textcolor{red}{568})}$ & $1\,243_{(751+\textcolor{red}{492})}$ & 2\,385                & 11h            & \textgreater 10  \\
$\textsc{LRS2-Common}_{\scriptsize\texttt{TF\textgreater   20}}$  & 17\,247          & $1\,082_{(389+\textcolor{red}{693})}$ & $1\,243_{(631+\textcolor{red}{612})}$ & 1\,413                & 8h             & \textgreater 20  \\
$\textsc{LRS2-Common}_{\scriptsize\texttt{TF\textgreater   50}}$  & 10\,122          & $1\,082_{(231+\textcolor{red}{851})}$ & $1\,243_{(416+\textcolor{red}{827})}$ & 626                 & 4h             & \textgreater 50  \\
$\textsc{LRS2-Common}_{\scriptsize\texttt{TF\textgreater   100}}$ & 5\,885           & $1\,082_{(135+\textcolor{red}{947})}$ & $1\,243_{(253+\textcolor{red}{990})}$ & 344                 & 2h             & \textgreater 100 \\
\bottomrule
\end{tabular}
\vspace{-1em}
\end{table*}

\section{Experiment}

\subsection{Datasets}

\paragraph{LRS2} \citet{afouras2018deep} is one of the most commonly used publicly available English wild lip-reading datasets, including 224 hours of video extracted from shows on BBC television. In the original dataset, the training data is divided into two partitions: \textit{Pretrain} (195H) and \textit{Train} (29H), both of which are transcribed from videos to text at the sentence level. The only difference is that the video clips in the \textit{Pretrain} partition is not strictly trimmed and sometimes longer than the corresponding text. 
We conducted experiments on LRS2 with different training data amounts (i.e., \textit{Pretrain+Train}(224h) and \textit{Train}(29h)).
Note that since the video of the LRS3 dataset has already been used in the pre-training process of AV-Hubert, we do not conduct experiments on it. 

\paragraph{LRS2-COMMON}
Based on the LRS2 dataset, we further propose the LRS2-COMMON to verify the lip-reading performance of the few-shot model trained with labeled visual utterances containing only common words. We counted the word frequency of each word in the \textit{Train} partition of LRS2, and extracted new training sets with only common words according to the word frequency. Note that during the inference, we use the complete \textit{Test} and \textit{Validation} that contain not only common words.


\subsection{Evaluation and Implementation Details}
For all experiments on LRS2, we use the \textit{word error rate} (WER) as the evaluation index of speech recognition (both lip-reading and AVSR). WER can be defined as $\texttt{WER}\textit{=}{(S+D+I)}/{M}$, where \(S,D,I,M\) represent the number of words replaced, deleted, inserted and referenced respectively. During validation, the inference is only performed when all of the validation utterances are of the same modality as the training utterances. For example, zero-shot lip-reading trained on labeled audio utterances should 
also be validated with audio utterances (the inference is performed on the visual utterances, during the testing). In Section \ref{sec:implementation}, we present more implement details.

\subsection{Main Result}

\begin{table}[tb]
\centering
\small
\tabcolsep=0pt
\caption{Comparison of full-shot, few-shot and zero-shot methods on LRS2.\protect\footnotemark[1] The experiments highlighted by \underline{the underline} are in the zero-shot setting.}
\begin{tabular}{ccccc}
\toprule

\multirow{2}{*}{\textbf{Type}}                                                                         & \multirow{2}{*}{\textbf{Method}} & \multicolumn{2}{c}{\textbf{Labeled Utt(hrs)}}                             & \multirow{2}{*}{\textbf{WER}(\%)}    \\ \cmidrule{3-4} 
                                                                                              &                         & \multicolumn{1}{c}{\textbf{Video}} & \multicolumn{1}{c}{\textbf{Audio}} &  \\ \midrule
\multirow{8}{*}{Full-Shot}                                                                   & \citet{son2017lip}      & 224                             & -                               & 70.4    \\
                                                                                              & \citet{afouras2018deep} & 698                             & -                               & 49.8    \\
                                                                                              & \citet{zhao2020hearing} & 698                             & 698                             & 65.3    \\
                                                                                              & \citet{zhang2019spatio} & 698                             & -                               & 51.7    \\
                                                                                              & \citet{afouras2020asr}  & 224                             & 808                             & 51.3    \\
                                                                                              & \citet{ren2021learning} & 698                             & 698                             & 49.2    \\
                                                                                              & \citet{prajwal2022sub}  & 698                             & -                               & 28.9    \\
                                                                                              & \citet{shi2022learning} & 224                             & -                               & 28.6    \\
                                                                                              & OpenSR(ours)  & 224                             & 224                               & \textbf{25.0}    \\
                                                                                              \midrule
\multirow{2}{*}{\begin{tabular}[c]{@{}c@{}}Few-Shot\end{tabular}} & \citet{afouras2020asr}        & 224                             & 1032                            & 54.2    \\
                                                                                              & \citet{ma2021lira}      & 224                             & -                               & 39.1    \\ \midrule
\multirow{2}{*}{\begin{tabular}[c]{@{}c@{}}Zero-Shot\end{tabular}}  & OpenSR(ours)                  & \textcolor{black}{\ding{56}}                               & 29                              &  \underline{39.2}       \\
                                                                                              & OpenSR(ours)                  & \textcolor{black}{\ding{56}}                               & 224                             & \underline{\textbf{36.0}}    \\ \bottomrule
\end{tabular}
\label{main_result}
\end{table}
\footnotetext[1]{The results presented are all trained using publicly available datasets.}

\begin{table*}[tb]
\small
\centering
\caption{Comparison of cross-domain method and cross-modality method in training lip-reading model for LRS2-BBC.  \begin{minipage}[b]{.085\linewidth}\includegraphics[scale=0.05]{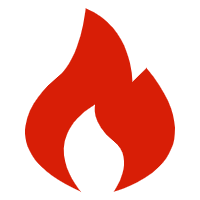} and \includegraphics[scale=0.05]{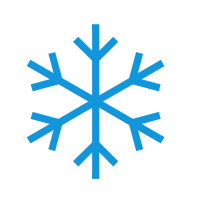}\end{minipage} indicate that the parameters in the Transformer Encoder are tunable and frozen, respectively. The experiments highlighted by \underline{the underline} are in the zero-shot setting. }
\begin{tabular}{cccccccc}
\toprule
\multirow{2}{*}{\textbf{Pretrain   Model}} & \multirow{2}{*}{\textbf{Training Split}}                                                & \multicolumn{3}{c}{\textbf{Training}}                 & \multirow{2}{*}{\textbf{\begin{tabular}[c]{@{}c@{}}Encoder\\      Frozen\end{tabular}}}                                                                                             & \multirow{2}{*}{\textbf{WER(\%)}} & \multirow{2}{*}{\textbf{ID}} \\ \cmidrule{3-5}
                                           &                                                                                         & \textbf{Utt(hrs)}           & \textbf{A} & \textbf{V} &                                                                                                                                                                                     &                                   &                     \\ \hline
\multirow{10}{*}{Transformer-BASE}         & \multirow{5}{*}{Train}                                                                  & LRS3(30h)                   & \textbf{}  & \ding{52}  & \begin{minipage}[b]{.1\linewidth}\centering\includegraphics[scale=0.05]{iclr2023/images/transfer.png}\end{minipage}                                                                 & 54.9                              & (1)                 \\ \cline{3-8} 
                                           &                                                                                         & \multirow{4}{*}{LRS2(29h)}  &            & \ding{52}  & \begin{minipage}[b]{.1\linewidth}\centering\includegraphics[scale=0.05]{iclr2023/images/transfer.png}\end{minipage}                                                                 & 51.2                              & (2)                 \\
                                           &                                                                                         &                             & \ding{52}  &            & \begin{minipage}[b]{.1\linewidth}\centering\includegraphics[scale=0.05]{iclr2023/images/transfer.png}\end{minipage}                                                                 & \underline{98.3}                              & (3)                 \\
                                           &                                                                                         &                             & \ding{52}  &            & \begin{minipage}[b]{.1\linewidth}\centering\includegraphics[scale=0.05]{iclr2023/images/frozen.png}\end{minipage}                                                                   & \underline{46.0}                              & (4)                 \\
                                           &                                                                                         &                             & \ding{52}  & \ding{52}  & \begin{minipage}[b]{.15\linewidth}\centering\includegraphics[scale=0.05]{iclr2023/images/frozen.png}(A)+\includegraphics[scale=0.05]{iclr2023/images/transfer.png}(V)\end{minipage} & \textbf{35.7}                     & (5)                 \\ \cline{2-8} 
                                           & \multirow{5}{*}{\begin{tabular}[c]{@{}c@{}}Train\\      +\\      Pretrain\end{tabular}} & LRS3(433h)                  &            & \ding{52}  & \begin{minipage}[b]{.1\linewidth}\centering\includegraphics[scale=0.05]{iclr2023/images/transfer.png}\end{minipage}                                                                 & 45.3                              & (6)                 \\ \cline{3-8} 
                                           &                                                                                         & \multirow{4}{*}{LRS2(224h)} &            & \ding{52}  & \begin{minipage}[b]{.1\linewidth}\centering\includegraphics[scale=0.05]{iclr2023/images/transfer.png}\end{minipage}                                                                 & 39.7                              & (7)                 \\
                                           &                                                                                         &                             & \ding{52}  &            & \begin{minipage}[b]{.1\linewidth}\centering\includegraphics[scale=0.05]{iclr2023/images/transfer.png}\end{minipage}                                                                 & \underline{98.5}                              & (8)                 \\
                                           &                                                                                         &                             & \ding{52}  &            & \begin{minipage}[b]{.1\linewidth}\centering\includegraphics[scale=0.05]{iclr2023/images/frozen.png}\end{minipage}                                                                   & \underline{42.3}                              & (9)                 \\
                                           &                                                                                         &                             & \ding{52}  & \ding{52}  & \begin{minipage}[b]{.15\linewidth}\centering\includegraphics[scale=0.05]{iclr2023/images/frozen.png}(A)+\includegraphics[scale=0.05]{iclr2023/images/transfer.png}(V)\end{minipage} & \textbf{31.7}                     & (10)                \\ \hline
\multirow{10}{*}{Transformer-LARGE}        & \multirow{5}{*}{Train}                                                                  & LRS3(30h)                   &            & \ding{52}  & \begin{minipage}[b]{.1\linewidth}\centering\includegraphics[scale=0.05]{iclr2023/images/transfer.png}\end{minipage}                                                                 & 43.0                              & (11)                \\ \cline{3-8} 
                                           &                                                                                         & \multirow{4}{*}{LRS2(29h)}  &            & \ding{52}  & \begin{minipage}[b]{.1\linewidth}\centering\includegraphics[scale=0.05]{iclr2023/images/transfer.png}\end{minipage}                                                                 & 31.4                              & (12)                \\
                                           &                                                                                         &                             & \ding{52}  &            & \begin{minipage}[b]{.1\linewidth}\centering\includegraphics[scale=0.05]{iclr2023/images/transfer.png}\end{minipage}                                                                 & \underline{98.2}                              & (13)                \\
                                           &                                                                                         &                             & \ding{52}  &            & \begin{minipage}[b]{.1\linewidth}\centering\includegraphics[scale=0.05]{iclr2023/images/frozen.png}\end{minipage}                                                                   & \underline{39.2}                              & (14)                \\
                                           &                                                                                         &                             & \ding{52}  & \ding{52}  & \begin{minipage}[b]{.15\linewidth}\centering\includegraphics[scale=0.05]{iclr2023/images/frozen.png}(A)+\includegraphics[scale=0.05]{iclr2023/images/transfer.png}(V)\end{minipage} & \textbf{29.5}                     & (15)                \\ \cline{2-8} 
                                           & \multirow{5}{*}{\begin{tabular}[c]{@{}c@{}}Train\\      +\\      Pretrain\end{tabular}} & LRS3(433h)                  &            & \ding{52}  & \begin{minipage}[b]{.1\linewidth}\centering\includegraphics[scale=0.05]{iclr2023/images/transfer.png}\end{minipage}                                                                 & 38.8                              & (16)                \\ \cline{3-8} 
                                           &                                                                                         & \multirow{4}{*}{LRS2(224h)} &            & \ding{52}  & \begin{minipage}[b]{.1\linewidth}\centering\includegraphics[scale=0.05]{iclr2023/images/transfer.png}\end{minipage}                                                                 & 28.6                              & (17)                \\
                                           &                                                                                         &                             & \ding{52}  &            & \begin{minipage}[b]{.1\linewidth}\centering\includegraphics[scale=0.05]{iclr2023/images/transfer.png}\end{minipage}                                                                 & \underline{97.4}                              & (18)                \\
                                           &                                                                                         &                             & \ding{52}  &            & \begin{minipage}[b]{.1\linewidth}\centering\includegraphics[scale=0.05]{iclr2023/images/frozen.png}\end{minipage}                                                                   & \underline{36.0}                              & (19)                \\
                                           &                                                                                         &                             & \ding{52}  & \ding{52}  & \begin{minipage}[b]{.15\linewidth}\centering\includegraphics[scale=0.05]{iclr2023/images/frozen.png}(A)+\includegraphics[scale=0.05]{iclr2023/images/transfer.png}(V)\end{minipage} & \textbf{25.0}                     & (20)                \\ \bottomrule
\end{tabular}
\label{cross}
\end{table*}

As shown in Table \ref{main_result}, we compare our method with the previous methods in LRS2 to highlight the effect of our proposed training system OpenSR. As the first training system that achieves zero-shot lip-reading, OpenSR not only achieves the state-of-the-art zero-shot and few-shot performance, but even outperforms most full-shot methods. This demonstrates that our training system can effectively train lip-reading models suitable for domains lacking labeled visual utterances. 
Furthermore, we demonstrate that OpenSR can improve the lip-reading capability of the full-shot lip-reading model. Since the features of audio and video are projected in the same phoneme space after pre-training, a decoder suitable for both ASR and VSR can be trained using only labeled audio utterances. Benefiting from this well-trained decoder, the performance is improved by 2.1\% compared to \cite{shi2022learning} using a similar framework.

\subsection{Cross Domain VS Cross Modality}
For the domains without labeled visual utterances, there are two ways to train lip-reading models using knowledge transfer: cross-domain and cross-modality. 
The experiments in Table \ref{cross} provide answers to the following questions about knowledge transfer:

\begin{itemize}[itemsep=-2pt,topsep = 0 pt]
    \item How much does the domain shift affect the lip-reading model?
    \item Is OpenSR training with cross-modality better than cross-domain transferring?
\end{itemize}

When there is no labeled visual utterances for lip-reading model training in the target domain (Here with LRS2), most of the current methods train on labeled visual utterances from other high resource domains (Here with LRS3). Compared with the model trained with in-domain utterances (ID: 2,7,12,17), the performance of models trained with other domains (ID: 1,6,11,16) utterances decreased by 3.7\% - 11.6\%, mainly because there may be terminological and syntactic differences among domains.
In Section \ref{sec:word_distribution}, we thoroughly discuss the domain differences between two datasets LRS2\&3 of the similar domain from the perspective of term distribution. 

In contrast, OpenSR training system learns the word distribution and syntax of the target domain from audio, which effectively avoids domain shift that affects the lip-reading model transferring between domains. As shown in Table \ref{cross}, even training without labeled visual utterances, the performance of OpenSR training with only audio utterances (ID: 4,9,14,19) can achieve considerable improvement (2.8\% - 8.9\%) in varying degrees comparing with cross-domain methods (ID: 1,6,11,16). In Section \ref{sec:qualitative}, we further present qualitative results on cross-domain and cross-modality.

\label{sec:cross}

\subsection{Pre-training alone cannot achieve zero-shot modality transfer.}
With the extraordinary performance of AV-Hubert on the uni-modality speech recognition, there might be some doubt that this zero-shot modality transfer benefits entirely from AV-Hubert. Indeed, pre-training does align different modalities into the phoneme space and enhance the feature representation, but as shown in the experiment (ID: 3,8,13,18) in Table \ref{cross}, the model with simple fine-tuning cannot achieve zero-shot modality transfer yet. The idea of parameter freezing in the second stage of OpenSR makes the model better maintain the alignment state between audio and visual features in phoneme space. Maintaining the multi-modality alignment is the key to zero-shot modality transfer, and OpenSR consistently achieves performance commensurate with the scale of the training utterances and model parameters with this property.

\begin{table}[h]
\centering
\small
\tabcolsep=2pt
\caption{Comparison of different strategies using only common words for training. `\textgreater N' indicates that the training visual utterances only contain words with TF\textgreater N in LRS2. \#Param is the tunable parameters during tuning.}
\begin{tabular}{lrcccc}
\toprule
                                             &                                     & \multicolumn{4}{c}{\scriptsize\textbf{WER in LRS2-COMMON}}                                                                       \\ \cmidrule{3-6} 
\multirow{-2}{*}{\textbf{Training   Strategy}} & \multirow{-2}{*}{\textbf{\# Param(MB)}} & \textbf{\textgreater 100} & \textbf{\textgreater 50} & \textbf{\textgreater 20} & \textbf{\textgreater 10} \\ 
\midrule
\citet{shi2022learning}     & 477.33($\times 1.00$)                            & 68.8                        & 61.8                       & 53.9                       & 49.6                       \\
{OpenSR}\\
\quad+Finetune             & 477.33($\times 1.00$)                            & 34.5                        & 33.2                       & 32.1                       & \textbf{30.3}         \\
\quad+Cluster prompt       & \textbf{9.84($\times$ 0.02)}                              & \textbf{32.7}               & \textbf{32.1}              & \textbf{30.8}              & 30.5                       \\ 
\bottomrule
\end{tabular}
\label{tab:common}
\end{table}

\subsection{Model Tuning with Common Word Videos}
\label{sec:prompt}
Table \ref{tab:common} compares the performance and tuning parameter scale with the state-of-the-art method in LRS2-COMMONs with different word frequencies. The domain shift seriously affects the training of the decoder in AV-Hubert, when the word frequency threshold of the tuning dataset (from $\textsc{LRS2-224H}$ to $\textsc{LRS2-COMMON}_{\scriptsize\texttt{TF\textgreater 100}}$) gradually increased, the WER of lip-reading increases sharply, from 28.6\% to 68.8\%.
By learning the word distribution and specific syntax in the target domain from audio, OpenSR ensures that the model will not overfit the common words even if only the video utterances containing a small number of common words is used, as shown in training strategy `OpenSR + Finetune'. 
Furthermore, our proposed Cluster Prompt training strategy in the few-shot scenario shows a further improvement compared to training strategy `OpenSR + Finetune' in terms of common words with a high word frequency threshold (`\textgreater 100', `\textgreater 50' and `\textgreater 20'). Also, note that the amount of the tuning parameters ($\times 0.02$) of the Cluster Prompt is significantly smaller than the amount ($\times 1.00$) in the other strategies. In particular, with the decrease of the frequency threshold of common words, the amount of tuning utterances keeps increasing, and the influence of domain shift gradually disappears. 
Compared to Cluster Prompt, the strategy of fine-tuning learns more about the distribution of the target domain that is applicable to lip-reading from the common word utterances, such as sequence modeling in the decoder specifically for the visual utterances. 


\begin{table}[tb]
\centering
\small
\tabcolsep=1pt
\caption{Ablation experiments of different tuning layers in the model. The encoder and decoder have 24 and 9 layers respectively.}
\begin{tabular}{lc}
\toprule
\textbf{Method}               &  \textbf{WER(\%)} \\ 
\midrule
OpenSR(zero-shot)             & 35.995           \\ 
\midrule
\textit{OpenSR+Tuning the Encoder Layers}\\
\quad + w/ encoder.layer.[18,24]             & 28.273           \\ 
\quad\quad + w/ encoder.layer.[12,18]             & 26.996           \\
\quad\quad\quad + w/ encoder.layer.[\,6,12]              & 26.711           \\
\quad\quad\quad\quad + w/ encoder.layer.[\,0,\,6]              & 26.426           \\
\midrule
\textit{OpenSR+Tuning the Decoder Layers}\\
\quad + w/ decoder.layer.[\,0,\,9]               & 32.522           \\ 
\midrule
\textit{OpenSR+Tuning the Encoder and Decoder Layers}\\
\quad + w/ encoder.layer.[\,0,24] \& decoder.layer.[\,0,\,9]               & 24.954           \\ 
\bottomrule
\end{tabular}
\label{ablation}
\end{table}

\subsection{Is the audio trained decoder suitable for lip reading?}

We conducted ablation experiments on modules participating in fine-tuning to explore why OpenSR could optimize the upper accuracy limit of the full-shot lip-reading, as shown in Table \ref{ablation}. Only 3.473\% (from 35.995\% to 32.522\%) improvement comes from tuning the decoder, while fine-tuning the pre-trained encoder can achieve additional 9.569\% improvement (from 35.995\% to 26.426\%). The limited improvement achieved by continuing to tune the decoder demonstrates that the word distribution and syntax learned in the audio modality can be zero-shot transferred to the visual modality. Meanwhile, this is why OpenSR can increase the accuracy ceiling, as the encoder pre-trained on large amounts of utterances and the decoder with knowledge of the target domain can steadily boost each other. Furthermore, we also attempt to determine the parameters that need to be tuned most during modality transfer.
When tuning the last few layers in the encoder, the performance boost is the most pronounced, increasing the performance by 7.722\% (from 35.995\% to 28.273\%), while additional tuning of all the other layers (encoder.layer.[0,18]) only provides a limited increase of 1.847\% (from 28.273\% to 26.426\%). This demonstrates that what needs to be further tuned while modality transfer is the target-modality representation of the encoder, so that it can distinguish visually similar words (e.g., \texttt{PET} and \texttt{BET}). By maintaining multi-modality feature alignment, the decoder trained with a single modality (i.e., audio) utterance can directly apply the knowledge of the target domain to lip-reading models.

\subsection{Modality Prompt VS Cluster Prompt}
\begin{figure}[tb]
    \centering
    \includegraphics[scale=0.47]{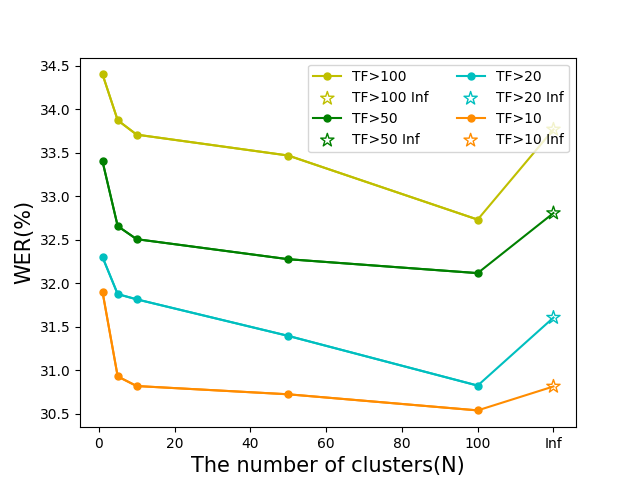}
    \caption{Comparison of lip-reading performance with different number of clusters(N). For comparison, the performance of $\inf$ in the figure is simply represented by N=500.}
    \label{fig:prompt}
\end{figure}

The Cluster Prompt enables the model to provide prompt embeddings according to the clusters of different phoneme features. In Figure \ref{fig:prompt}, we show the influence of the number of clusters on the performance of accuracy. Obviously, when there is only one cluster, it can be considered as providing the prompt embedding only for the visual modality, which can also be called Modality Prompt. With the increase of the number of clusters (N), the cluster division of the model becomes more refined, so that the prompt embedding also becomes more consistent with each representation. In particularly, we noticed that the accuracy decreases when the number of clusters increases to $\inf$ (the performance of $\inf$ in the figure is simply represented by n=500). This is because there is no enough data to support the training of over-differentiated cluster specific prompt embedding.

\subsection{Performance of AV Speech Recognition.}
OpenSR can also achieve zero-shot modality transfer from a single modality to multi-modality, by maintaining both audio and visual modality features aligned at the same time. As shown in Table \ref{tab:avsr}, OpenSR outperforms other methods in audio-visual speech recognition. Note that, even in the zero-shot setting, OpenSR performes better than the full-shot performance previous of most previous methods. Furthermore, only using 29h visual utterance to participate in training is enough to comprehensively surpass the previous methods. Comparing with AV-Hubert of the same backbone, the performance of OpenSR is 0.4\% better, demonstrating the significance of the term distribution and syntax specification learned during a single modality training for full-shot speech recognition in other-modality.


\begin{table}[h]
\small
\centering
\caption{Comparison of audio-visual speech recognition performance on LRS2. The experiment highlighted by \underline{the underline} is in the zero-shot setting.}
\begin{tabular}{cccc}

\toprule
                                  & \multicolumn{2}{c}{\textbf{Labeled Utt(hrs)}} &                                \\ 
\cmidrule{2-3}
\multirow{-2}{*}{\textbf{Method}} & \textbf{Video}       & \textbf{Audio}       & \multirow{-2}{*}{\textbf{WER(\%)}} \\ 
\midrule
\citet{afouras2018deep}           & 1428                 & 1428                 & 8.5                            \\
\citet{petridis2018audio}         & 381                  & 381                  & 7.0                            \\
\citet{yu2020audio}               & 224                  & 224                  & 5.9                            \\
\citet{ma2021lira}                & 224                  & 224                  & 3.7                            \\
\citet{shi2022learning}           & 224                  & 224                  & 3.1                         \\
\midrule
                                  & -                    & 224                  & \underline{3.3}                          \\
                                  & 29                   & 224                  & 2.8                          \\
\multirow{-3}{*}{OpenSR(ours)}    & 224                  & 224                  & \textbf{2.7}                 \\ 
\bottomrule
\end{tabular}
\label{tab:avsr}
\end{table}


\section{Conclusion}
The bottleneck of domain-specific models is the lack of target domain data, such as the challenge of visual utterance collection in lip-reading. We propose OpenSR, a training system that can train target domain lip-reading models without using labeled visual utterance. 
The other modality models can directly utilize the target domain knowledge obtained from the single modality (e.g., audio) utterance, via maintaining the multi-modality feature alignment state learned during pre-training. This zero-shot modality transfer idea can alleviate the training problem caused by the severely scarce labeled data of some modalities. For example, despite there is no labeled visual utterance of the target domain, we can still use OpenSR to train the lip-reading model with the labeled audio utterance for the target-domain specifically. 
Furthermore, our training system achieves zero-shot modality transfer in a tuning-based manner, bringing a new perspective utilizing the pre-trained models, which can be transferred to other multi-modality pre-training models such as CLIP \cite{Radford2021LearningTV}.

\section{Ethical Discussion}
Lip-reading has many applications, including instruction dictation in public areas or a noisy environment and information accessibility for the hearing impaired. OpenSR makes it possible to quickly build domain-specific lip-reading models for low-resource domains (lack of labeled visual utterances), which greatly enhances the fairness of lip-reading technology across the domains and languages.

For speech recognition, there may be concerns about the risk of information leakage. But in fact, as mentioned above, the lip-reading model has a relatively high requirement on the visual utterances used for lip-reading, requiring mostly-frontal and high-resolution videos with a sufficiently high frame rate, such that motions around the lip area are clearly captured. In general, only close-range cameras or online meetings have similar video conditions, which ensures that the lip-reading model will not be abused in potentially privacy revealing scenarios such as surveillance videos.

\section*{Acknowledgements}
This work was supported in part by the National Key R\&D Program of China under Grant No.2022ZD0162000,National Natural Science Foundation of China under Grant No. 62222211, Grant No.61836002 and Grant No.62072397, and Yiwise.
 

\bibliography{anthology,custom}
\bibliographystyle{acl_natbib}

\appendix
\newpage

\section{Implementation Details}
\label{sec:implementation}

\paragraph{Audio and Visual Utterance Preprocessing.}
For the visual utterance, we only intercept the lip region for lip-reading. As the previous methods \cite{shi2022learning,afouras2018deep,afouras2018lrs3}, we adopt dlib \cite{king2009dlib} to detect the 68 facial keypoints and align each face with its neighbors. We crop a $96\times96$ region-of-interest (ROI) talking head video centered on the mouth from each visual utterance.
For the audio utterance, we also remain same preprocessing steps as the prior works \cite{ma2021lira,shi2022learning}. We extract the 26-dimensional log filterbank energy feature from the raw waveform and stack the 4 neighboring acoustic frames for synchronization. During training, for data enhancement, we randomly crop $88\times88$ from the whole ROI and flipped it horizontally with 0.5 probability. To improve noise robustness, we apply noise with a probability of 0.25 to each audio utterance from \cite{Snyder2015MUSANAM} as steps in \cite{afouras2018deep}.

\paragraph{Pre-training Setup.}
OpenSR builds on pretraining process of AV-Hubert \cite{shi2022learning}, directly utilizing its checkpoint for the subsequent stages. During pre-training, a modified ResNet-18 used in prior works \cite{ma2021lira,martinez-2021-fujitsu} and a linear projection layer are adopted as visual and audio encoders, respectively. It considers two models with different configurations: Transformer-BASE and Transformer-LARGE have 12/24 Transformer layers with the embedding dimension/feed-forward dimension/attention heads of 768/3072/12 and 1024/4096/16. We simply adopted the pre-trained model obtained by training on LRS3 \cite{afouras2018lrs3} and VoxCeleb2 \cite{Chung2018VoxCeleb2DS}.

\paragraph{OpenSR Tuning Setup in a Single Modality.}
In the second stage of OpenSR, we fine-tune the decoder with the labeled audio utterance and the absent modality (visual) feature is replaced by a zero-vector. For comparison, we adopt the same decoder configuration as \cite{shi2022learning}, with 6 and 9 Transformer layers in Transformer-BASE and Transformer-LARGE, respectively.
With the encoder parameters frozen, we fine-tune the decoder on a single 3090 GPU for 45K/120K steps in the 29h/224h setting. Note that, also only audio can be used while tuning the hyperparamters on the validation set during the second stage.
Conversely, during the inference or further tuning with the visual utterance (the third stage), we only adopt the visual 
utterance as input and replace the audio feature with the zero-vector $\mathbf{0_D}$. Each stage of OpenSR is trained with Adam, with the learning rate being warmed up for the first 50\% of updates to 0.0005.

\begin{figure}[b]
    \centering
    \includegraphics[scale=0.4]{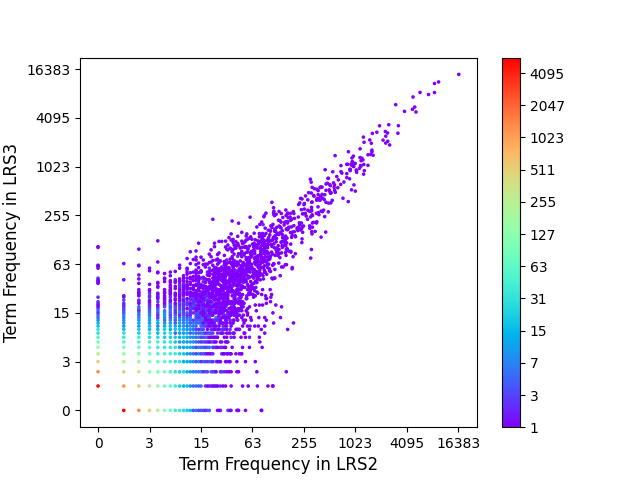}
    \caption{Different frequencies of the same word in LRS2\&3. The color shows the number of words at that position.}
    \label{fig:fre}
\end{figure}

\section{Word distribution differences between domains}
\label{sec:word_distribution}

In this section, starting with LRS2 and LRS3, we explore the differences in word distribution in different domains. Although both datasets are extracted from television shows (BBC and TED respectively), we still find a huge difference in word distribution between them, perhaps due to the different content of the shows.

\begin{figure}[t]
    \centering
    \includegraphics[scale=0.4]{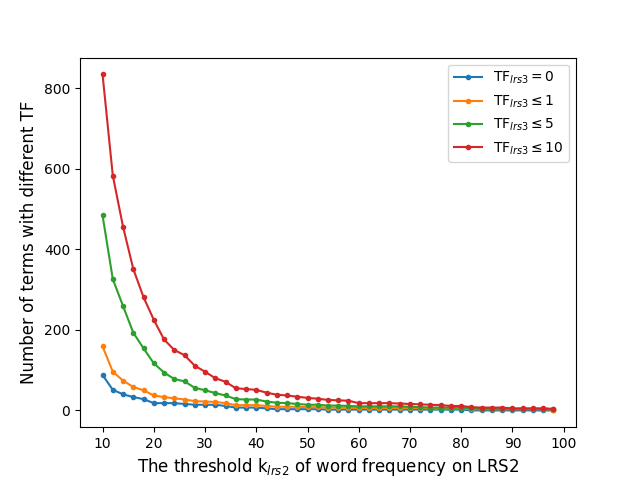}
    \caption{The number of terms whose $\textsc{TF}_{lrs2} \geq \textsc{k}_{lrs2}$ and $\textsc{TF}_{lrs3}\leq \textsc{k}_{lrs3}$. $\textsc{TF}_{lrs2\&3}$ and $\textsc{k}_{lrs2\&3}$ are the word frequency of each word and the counting threshold on LRS2\&3, respectively.}
    \label{fig:missing}
\end{figure}

\paragraph{Different frequencies of the same word in different domains.}
In Figure \ref{fig:fre}, we visually show the word frequency of each word in LRS2\&3, each domain has more than 4,000 words that do not appear in the other's dictionary, as shown by the red dots. There are also a number of words that vary widely in word frequency from domain to domain, as shown by the points off the diagonal. In general, words that are far off the diagonal tend to be terms specific to the domain. Except for a few words with high frequencies that are common to all domains, most words have different word frequencies in different domains. From the perspective of common terms in LRS2, we further quantitatively reveal the differences of the common terms distribution in Figure \ref{fig:missing}. 
There are 834 words (2.02\% of the LRS2 vocabulary size) with word frequency differences (greater than and less than 10, respectively), demonstrating that a large part of the terms is domain-specific. At the same time, there are a number of terms that varied greatly between domains: 88 words with frequency over 10 (e.g., \textit{bargain}, \textit{crafts} and \textit{saxon}) and 2 words with frequency over 80 (\textit{cos} and \textit{antiques}) never appears in LRS3.

\begin{figure}[t]
    \centering
    \includegraphics[scale=0.4]{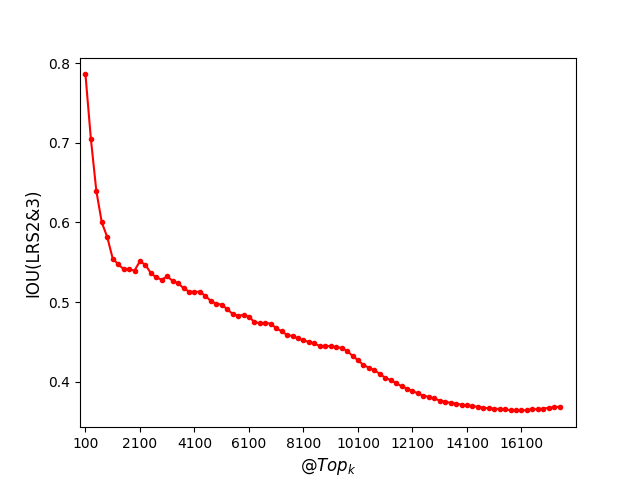}
    \caption{The IOU of the words with the highest @$Top_k$ word frequency between LRS2 and LRS3.}
    \label{fig:iou}
\end{figure}
\paragraph{The IoU of words in LRS2 and LRS3}
In Figure \ref{fig:iou}, we show the IoU (Intersection over Union) of the words with the highest @$Top_k$ word frequency in two datasets. Among the top 100 words in the two datasets, there are still 21 different words in addition to non-domain-specific generic words such as `The', `A' and `I' et al. From the perspective of the whole dataset, the IoU of the dictionaries in the two datasets is only 36.925\%, which means that there are a large number of words that occur only in their respective domains. Even between near-domain datasets LRS2 and LRS3, there are differences both in the most commonly used words and in the whole dictionary. In certain domains, such as biomedicine and electronic information, the greater difference in word distribution between domains makes the lip-reading model unable to transfer across domains.

\begin{table}[t]
    \centering
    \caption{Video clips of visually confusable words.}
    \begin{tabular}{l}
    \toprule
        Reference: in touch\\
        confusable answer: searched\\
        \begin{minipage}[b]{0.90\linewidth}\includegraphics[scale=0.50]{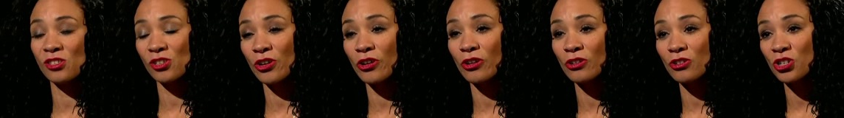}\end{minipage} \\ \hline
        Reference: rectify\\
        confusable answer: ratify \\
        \begin{minipage}[b]{0.90\linewidth}\includegraphics[scale=0.50]{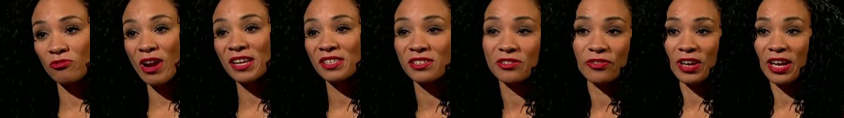}\end{minipage} \\ \hline        
        Reference: britain's\\
        confusable answer: prints \& print's\\
        \begin{minipage}[b]{0.90\linewidth}\includegraphics[scale=0.50]{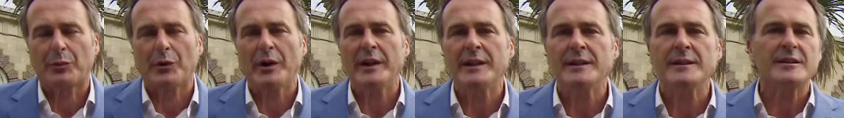}\end{minipage} \\ \hline
        Reference: fingers\\
        confusable answer: fears \& feelers\\
        \begin{minipage}[b]{0.90\linewidth}\includegraphics[scale=0.50]{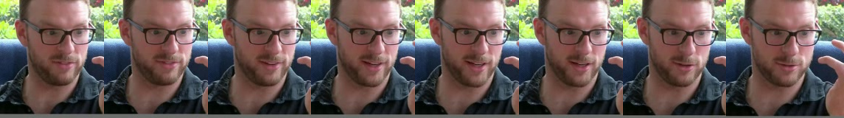}\end{minipage} \\
    \bottomrule
    \end{tabular}
    \label{tab:confusable}
\end{table}

\section{Qualitative Results}
\label{sec:qualitative}

\paragraph{Performance on visually confusing words.}
Zero-shot OpenSR trained using only audio performed worse on similar-sounding words. We show several video clips of visually confusable words in Table \ref{tab:confusable}, and show the performance comparison of the model trained with different scale visual utterances in Table \ref{fig:audio}. With the gradual introduction of visual utterances, the lip-reading performance of the model for visually confusable words is significantly enhanced, demonstrating the significance of our proposed further training using the utterances of common words.

\paragraph{Performance with different term distributions}
The distribution of terms will seriously affect the recognition results, while the training speech recognition models. If a term appears never or only a few times, it is essentially unrecognized, as shown in the Table \ref{fig:domain}. We notice that the models are more likely to come up more common answers (words with high frequency in the in-domain dictionary), probably because the model needs to be trained to fit the word distribution in the training utterances.



\renewcommand{\floatpagefraction}{1}
\begin{table*}
\caption{Qualitative comparison on visually confusing words. \textcolor{red}{Red words} highlights misidentified words, \textcolor{red}{(\st{strikeouts})} in parentheses highlight corresponding visually similar words and the \textcolor{red}{(red words)} in parentheses highlight the absent words.}
\centering
\tabcolsep=2pt
\begin{tabular}{ll}
\toprule
\multicolumn{2}{l}{\begin{minipage}[b]{\linewidth}\centering\includegraphics[scale=0.75]{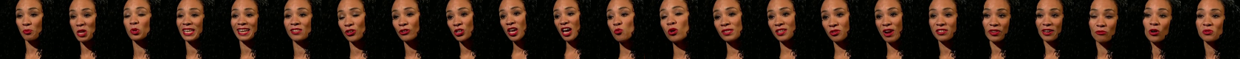}\end{minipage}} \\ 
\midrule
Ground Truth:                                           & people getting in touch and wanting to rectify wrongs                                 \\
$\textrm{OpenSR}_{\scriptsize{\textrm{Zero-shot}}}$(0h):             & people getting \textcolor{red}{searched (\st{in touch})} and wanting to \textcolor{red}{ratify (\st{rectify}) what was}           \\
$\textrm{OpenSR}_{\scriptsize{\textrm{Few-shot}}}$(2h):              & people getting in touch and wanting to rectify \textcolor{red}{what is}                                \\
$\textrm{OpenSR}_{\scriptsize{\textrm{Full-shot}}}$(433h):           & people getting in touch and wanting to rectify \textcolor{red}{what's}                                 \\ 
\midrule
\midrule
\multicolumn{2}{l}{\begin{minipage}[b]{\linewidth}\centering\includegraphics[scale=0.75]{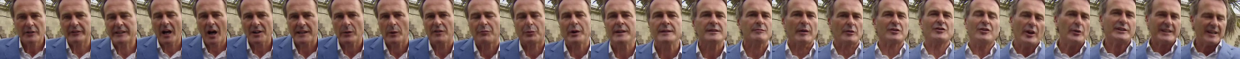}\end{minipage}} \\ 
\midrule
Ground Truth:                                                           & regarding one of britain's most noted                                 \\
$\textrm{OpenSR}_{\scriptsize{\textrm{Zero-shot}}}$(0h):                & regarding one of \textcolor{red}{prints (\st{britain's})} most \textcolor{red}{noting}           \\
$\textrm{OpenSR}_{\scriptsize{\textrm{Few-shot}}}$(2h):                 & regarding one of \textcolor{red}{print’s (\st{britain's})} most \textcolor{red}{noting}                                \\
$\textrm{OpenSR}_{\scriptsize{\textrm{Full-shot}}}$(433h):              & regarding one of britain's most \textcolor{red}{noting}                                 \\ 
\midrule
\midrule
\multicolumn{2}{l}{\begin{minipage}[b]{\linewidth}\centering\includegraphics[scale=0.75]{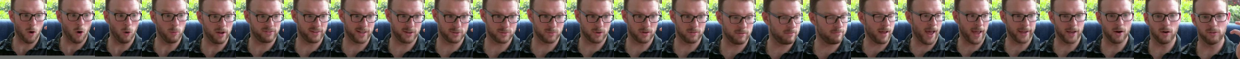}\end{minipage}} \\ 
\midrule
Ground Truth:                                           & all my fingers were hanging off                                                       \\
$\textrm{OpenSR}_{\scriptsize{\textrm{Zero-shot}}}$(0h):             & all my \textcolor{red}{fears (\st{fingers})} were hanging \textcolor{red}{(off)}                                               \\
$\textrm{OpenSR}_{\scriptsize{\textrm{Few-shot}}}$(2h):              & all my \textcolor{red}{feelers (\st{fingers})} were hanging off                                             \\
$\textrm{OpenSR}_{\scriptsize{\textrm{Full-shot}}}$(433h):           & all my fingers were hanging off                                                       \\ 
\midrule
\bottomrule
\end{tabular}
\label{fig:audio}
\end{table*}

\begin{table*}
\centering
\tabcolsep=1pt
\caption{Qualitative performance comparison of models trained with utterances of different word distribution. The misidentified words are highlighted in \textcolor{red}{red} and the absent words are highlighted with \textcolor{red}{(red)} in parentheses. The table on the right shows the word frequency shift of the misidentified words in different domains.}
\begin{tabular}{lllccc}

\toprule
\multicolumn{2}{l}{\begin{minipage}[b]{0.65\linewidth}\centering\includegraphics[scale=0.65]{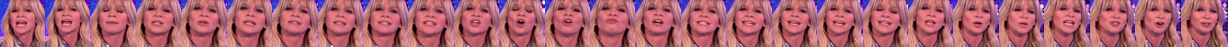}\end{minipage}} &  & \multicolumn{3}{c}{\small{\textbf{Word Frequency}}}                       \\ \cline{1-2} \cline{4-6} 
\small{Ground Truth}:                                   & he absolutely insisted on writing his own intro         &  & \multirow{2}{*}{\textbf{Term}} & \multicolumn{2}{c}{\textbf{Dataset}}        \\ \cline{5-6} 
\small{AV-Hubert}\small{(LRS3)}:                                & \textcolor{red}{(he)} absolutely \textcolor{red}{assisted} on writing his own intro         &  &                                & \scriptsize{\textbf{LRS2}}        & \scriptsize{\textbf{LRS3}}        \\ \cline{4-6} 
$\textrm{\small{OpenSR}}_{\scriptsize{\textrm{zero-shot}}}$\small{(LRS2)}:   & he absolutely insisted on writing his own \textcolor{red}{entro}         &  & insisted                       & 2                    & 0                    \\
$\textrm{\small{OpenSR}}_{\scriptsize{\textrm{full-shot}}}$\small{(LRS2)}:   & he absolutely insisted on writing his own \textcolor{red}{injury}        &  & intro                          & 0                    & 1                    \\ 

\cline{1-2} \cline{4-6} 
\midrule

\multicolumn{2}{l}{\begin{minipage}[b]{0.65\linewidth}\centering\includegraphics[scale=0.65]{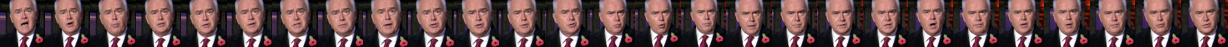}\end{minipage}} &  & \multicolumn{3}{c}{\small{\textbf{Word Frequency}}}                        \\ \cline{1-2} \cline{4-6} 
\small{Ground Truth}:                                   & an opportunity to pay tribute to all members of the armed forces               &  & \multirow{2}{*}{\textbf{Term}} & \multicolumn{2}{c}{\textbf{Dataset}}        \\ \cline{5-6} 
\small{AV-Hubert}\small{(LRS3)}:                                & an opportunity to pay \textcolor{red}{attribute} to all members of the armed forces             &  &                                & \scriptsize{\textbf{LRS2}}        & \scriptsize{\textbf{LRS3}}        \\ \cline{4-6} 
$\textrm{\small{OpenSR}}_{\scriptsize{\textrm{zero-shot}}}$\small{(LRS2)}:   & an opportunity to pay tribute to all members of the armed forces               &  & tribute                        & 5                    & 0                    \\
$\textrm{\small{OpenSR}}_{\scriptsize{\textrm{full-shot}}}$\small{(LRS2)}:   & an opportunity to pay tribute to all members of the armed forces               &  & attribute                      & 0                    & 3                    \\ 

\cline{1-2} \cline{4-6}  
\midrule

\multicolumn{2}{l}{\begin{minipage}[b]{0.65\linewidth}\centering\includegraphics[scale=0.65]{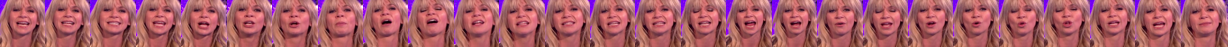}\end{minipage}} &  & \multicolumn{3}{c}{\small{\textbf{Word Frequency}}}                     \\ 
\cmidrule{1-2} \cmidrule{4-6} 
\small{Ground Truth}:                                   & let's take a look behind the scenes at how it all came         &  & \multirow{2}{*}{\textbf{Term}} & \multicolumn{2}{c}{\textbf{Dataset}}        \\ \cline{5-6} 
\small{AV-Hubert}\small{(LRS3)}:                                & let's take a look behind the \textcolor{red}{seeds (at)} how it all came         &  &                                & \scriptsize{\textbf{LRS2}}        & \scriptsize{\textbf{LRS3}}        \\ \cline{4-6} 
$\textrm{\small{OpenSR}}_{\scriptsize{\textrm{zero-shot}}}$\small{(LRS2)}:   & let's take a look behind the scenes \textcolor{red}{(at)} how it all came         &  & scenes                         & 14                   & 2                    \\
$\textrm{\small{OpenSR}}_{\scriptsize{\textrm{full-shot}}}$\small{(LRS2)}:   & let's take a look behind the scenes \textcolor{red}{(at)} how it all game        &  & seeds                          & 3                    & 7                    \\ 
\cline{1-2} \cline{4-6} 
\bottomrule
\end{tabular}
\label{fig:domain}
\end{table*}

\begin{table*}
\centering
\caption{Comparison of Lip-Reading performance on LRS3. VC2-EN stands for the English utterances of VoxCeleb2. For unlabelled utterances, only audio-visual speech can be employed for training (no corresponding transcription). Experiments labeled with $\dagger$ used non publicly available dataset.}
\begin{tabular}{clcrc}
\toprule
\multirow{2}{*}{\textbf{Mode}} & \multirow{2}{*}{\textbf{Method}} & \multirow{2}{*}{\textbf{Unlabeled Utts}} & {\textbf{Labeled Utts}} & \multirow{2}{*}{\textbf{WER(\%)}} \\ 
                              &                                  &                                               & \textbf{Video(hrs)}                              &                      \\ \midrule
\multirow{7}{*}{Full-Shot}     & Lira~\cite{ma2021lira}           & -                                                                     & 590                 & 43.3                 \\
                              & VisualSR~\cite{Ma2022VisualSR}   & -                                                                      & 1\,459                & 31.5                 \\
                              & $\dagger$~Sub~\cite{prajwal2022sub}         & -                                                                     & 2\,676                & 30.7                 \\
                              & $\dagger$~RecurrentNN~\cite{Makino2019RecurrentNN}    & -                                                                     & 31\,000               & 33.6                 \\
                              & $\dagger$~AV-VIT~\cite{Serdyuk2021AudioVisualSR} & -                                                                     & 90\,000               & \textbf{25.9}                 \\
                              \cmidrule{2-5} 
                              & AV-Hubert~\cite{shi2022learning}  & LRS3+VC2-EN                                                           & 433                 & 28.6                 \\
                              & OpenSR(ours)                     & LRS3+VC2-EN                                                          & 433                 & \textbf{28.5}                     \\ \midrule
Zero-Shot                      & OpenSR(ours)                     & LRS3+VC2-EN                                                           & \ding{55}                 & \textbf{30.6}                 \\ \bottomrule
\end{tabular}
\label{tab:lrs3}
\end{table*}

\section{Zero-Shot Lip-Reading on LRS3.}
We further present the performance on the LRS3 dataset in Table \ref{tab:lrs3} (although it has been used in pre-training). During pre-training, the encoder has fully mastered the domain-specific knowledge (the word distribution and syntax) in LRS3, resulting in the performance of 30.6\% under the zero-shot setting that is very close to the performance of 28.5\% under full-shot (+2.1\%). Note that, on the LRS2 dataset, which is not used during the pretraining, the performance of zero-shot is 11\% worse than that of full-shot (36.0\%->25.0\%). In fact, the difference between the further tuning effects on LRS2 and LRS3 (11\% and 2.1\%) effectively can also effectively demonstrate the domain shift between the LRS2 and LRS3 datasets.

\end{document}